\documentclass{preprint}

\usepackage{lipsum} 
\usepackage{xcolor}
\usepackage{booktabs}
\usepackage{graphicx} 
\usepackage[numbers,comma,sort&compress,super]{natbib}
\usepackage[colorlinks=true]{hyperref}
\usepackage[para,online,flushleft]{threeparttable}
\usepackage{amsmath}
\usepackage{pdfpages}

\title{Harnessing the power of longitudinal medical imaging for eye disease prognosis using Transformer-based sequence modeling}
\author[1,2]{Gregory Holste}
\author[2]{Mingquan Lin}
\author[3]{Ruiwen Zhou}
\author[1]{Fei Wang}
\author[3]{Lei Liu}
\author[4]{Qi Yan}
\author[5]{Sarah H. Van Tassel}
\author[5]{Kyle Kovacs}
\author[6]{Emily Y. Chew}
\author[7]{Zhiyong Lu}
\author[2,*]{Zhangyang Wang}
\author[1,*]{Yifan Peng}
\affil[1]{Department of Population Health Sciences, Weill Cornell Medicine, NY, USA}
\affil[2]{Department of Electrical and Computer Engineering, The University of Texas at Austin, TX, USA}
\affil[3]{Center for Biostatistics and Data Science, Washington University School of Medicine, St. Louis, MO, USA}
\affil[4]{Department of Obstetrics \& Gynecology, Columbia University, New York, NY, USA}
\affil[5]{Department of Ophthalmology, Weill Cornell Medicine, New York, USA}
\affil[6]{Division of Epidemiology and Clinical Applications, National Eye Institute, National Institutes of Health (NIH), Bethesda, MD, USA}
\affil[7]{National Center for Biotechnology Information (NCBI), National Library of Medicine (NLM), National Institutes of Health (NIH), Bethesda, MD, USA}
\affil[*]{Corresponding: \url{yip4002@med.cornell.edu} and \url{atlaswang@utexas.edu}}

\begin{document}

\maketitle

\begin{abstract}
Deep learning has enabled breakthroughs in automated diagnosis from medical imaging, with many successful applications in ophthalmology. However, standard medical image classification approaches only assess disease presence \emph{at the time of acquisition}, neglecting the common clinical setting of longitudinal imaging. For slow, progressive eye diseases like age-related macular degeneration (AMD) and primary open-angle glaucoma (POAG), patients undergo repeated imaging over time to track disease progression and forecasting the future risk of developing a disease is critical to properly plan treatment. Our proposed Longitudinal Transformer for Survival Analysis (LTSA) enables dynamic disease prognosis from longitudinal medical imaging, modeling the time to disease from sequences of fundus photography images captured over long, irregular time periods. Using longitudinal imaging data from the Age-Related Eye Disease Study (AREDS) and Ocular Hypertension Treatment Study (OHTS), LTSA significantly outperformed a single-image baseline in 19/20 head-to-head comparisons on late AMD prognosis and 18/20 comparisons on POAG prognosis. A temporal attention analysis also suggested that, while the most recent image is typically the most influential, prior imaging still provides additional prognostic value.
\end{abstract}


\section{Introduction}

Deep learning has shown remarkable capabilities across a wide variety of medical image analysis and computer-aided diagnosis tasks,\cite{Esteva2021-wd, Chen2022-jq} with many successful applications in ophthalmology.\cite{Lin2023-hv, Chen2021-Multimodal, Peng2019-ee, Chen2019-pe} 
However, standard image classification techniques for disease diagnosis bear some major limitations: (i) they can only accommodate a single image of a patient, and (ii) they can only assess if the patient presents with the disease at the time of image acquisition. For slowly progressive eye diseases like late-stage age-related macular degeneration (late AMD) and primary open-angle glaucoma (POAG), it is common for patients to undergo repeated imaging over long periods of time to track disease progression. Such longitudinal imaging is incompatible with standard image classification methods, though a growing body of work tackles this common clinical setting.\cite{Peng2020-pp, Lee2022-sq, Ghahramani2021-ek, Yan2020-ou, Cascarano2023-na}
In addition, for patients who do not present with the disease at the time of acquisition but may be at increased risk of developing it in the next few years, it is critical to identify this risk as early as possible to plan management. Further, patients in different subphenotypes might have varying eye disease progression speeds in earlier and later stages; long-term epidemiological studies have shown that many factors ``dynamically'' influence AMD or POAG progression.\cite{Ederer1994-vp, Miglior2005-bz}

For these reasons, we aim to develop a method for disease \emph{prognosis}, forecasting future risk of developing disease based on longitudinal imaging. Prior work has used color fundus photography imaging for AMD\cite{Peng2020-pp, Lee2022-sq, Ghahramani2021-ek, Yan2020-ou} and POAG\cite{Li2022-uk, Lin2022-Multi} prognosis, some incorporating prior imaging. For example, Peng et al.\cite{Peng2020-pp} adopted a two-stage approach where a convolutional neural network (CNN) was pretrained on AMD-related tasks to be used as a fundus image feature extractor. Next, these embeddings were used alongside patient demographics and genomic data in a Cox proportional hazards model for survival analysis modeling of ``time to late AMD'' from the last available image. Yan et al.\cite{Yan2020-ou} developed an end-to-end deep learning approach for late AMD progression by training a CNN to predict the risk of developing late AMD over $k$ years, where $k = 2,3,\ldots,7$, again based on individual fundus images. More recently, Lee et al.\cite{Lee2022-sq} and Ghahramani et al.\cite{Ghahramani2021-ek} began to incorporate longitudinal fundus imaging for automated late AMD progression with a CNN feature extractor and separate long short-term memory (LSTM) module to model temporal patterns in the imaging. Lee et al. classified whether the eye would develop late AMD in fixed windows of 2 or 5 years from the last image, similar to Yan et al., while Ghahramani et al. employed a two-stage approach with survival modeling based on deep fundus image features, much like Peng et al. Similarly, for POAG prognosis, Li et al.\cite{Li2022-uk} adopted a two-stage approach that extracts deep features from a baseline fundus image and performs survival modeling. Later, Lin et al.\cite{Lin2022-Multi} used a siamese CNN architecture to model changes between a baseline and follow-up fundus image, roughly modeling progression by classifying whether the eye would develop POAG within 2 and 5 years from the last visit.

Overall, these prior efforts often formulate automated prognosis as a binary classification task, for example, predicting whether a patient will develop the disease within fixed durations from the last visit (e.g., 2-year or 5-year prognosis). As a result, these approaches are limited to the time horizon of choice. Converting a 5-year risk classifier to a 2-year risk classifier would potentially entail creating a new patient cohort and re-training the model from scratch on this new classification task. Additionally, this model can only provide a single scalar describing the probability of developing disease at any time within the specified time window, unable to produce a time-varying risk assessment. Finally, many of these methods are only capable of accommodating a single fundus image, failing to model temporal changes in the eye's presentation that may be crucial for assessing the rate of progression and, thus, proper prognosis. To avoid these pitfalls, we adopt a survival analysis approach to disease prognosis from longitudinal imaging data, aiming to model a time-to-event outcome (e.g., years until death or developing a disease) based on time-varying patient measurements. Such an approach is far more flexible and clinically valuable than prior efforts toward eye disease prognosis, as it incorporates longitudinal patient imaging and produces dynamic and long-term risk assessments. For example, this approach would be particularly informative for a patient who has already ``survived'' several years with no current signs of disease or an early stage of eye disease. Further, our method is end-to-end, meaning it directly accepts longitudinal fundus images and outputs time-varying probabilities describing the risk of developing the disease of interest.

In this work, we propose a Longitudinal Transformer for Survival Analysis (LTSA), a Transformer-based method for end-to-end survival analysis based on longitudinal imaging. Like words in a sentence, we represent the collection of longitudinal images over time as a \emph{sequence} fit for modeling with Transformers.\cite{Vaswani2017-hz}
However, unlike words in a sentence, consecutive images are not ``equally spaced,'' potentially with months or years between visits. To account for this, a temporal positional encoder embeds the acquisition time (time elapsed since the first visit) of each image and fuses this information with the learned image embedding. A Transformer encoder then performs repeated ``causal'' masked self-attention operations, learning associations between the image from each visit and all prior imaging. The model is optimized to directly predict the discrete hazard distribution, a fundamental object of interest in survival analysis, from which we can construct eye-specific survival curves. Despite advances in deep learning for survival analysis,\cite{Wiegrebe2023-lp, Wolf2022-vw, Zhang2020-km, Chen2021-ba, Agarwal2021-oa, Katzman2018-pm, Ching2018-vs} existing methods either accommodate non-longitudinal imaging or non-imaging longitudinal (time series) data. LTSA is unique in its ability to perform end-to-end time-varying image representation learning and survival modeling from longitudinal imaging data using Transformer-based sequential modeling.

\begin{figure}
    \centering
    \includegraphics[width=\linewidth]{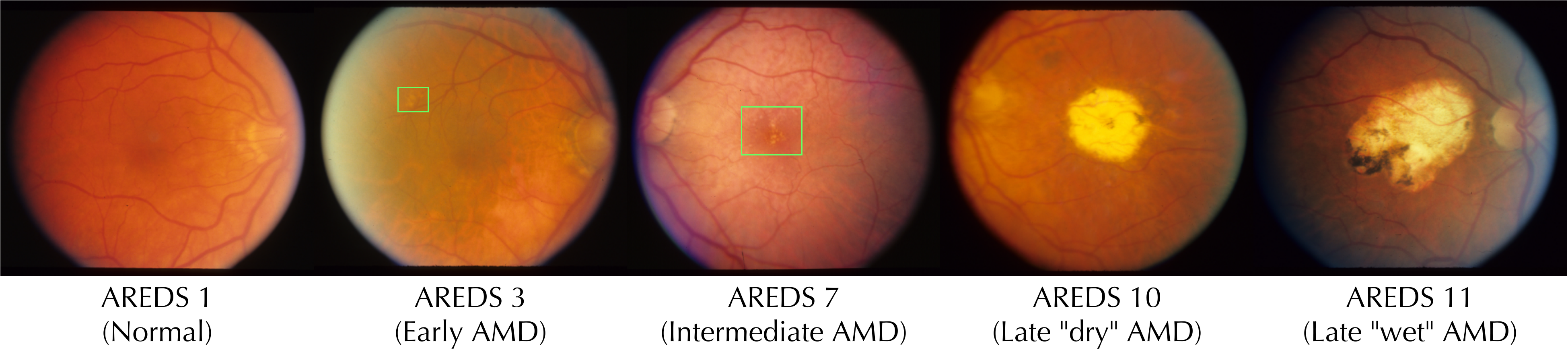}
    \caption{\textbf{Stages of AMD progression.} Color fundus photography images illustrating the various stages of AMD, a progressive eye disease affecting the macula. Images come from the AREDS dataset accompanied by a 9-step AMD severity score; a score over 9 indicates late-stage AMD, which can cause blurring and loss of central vision. Green boxes highlight ``drusen'', yellowish deposits of protein under the retina, which can be an early sign of AMD. There are two forms of late AMD: ``dry'', or atrophic, AMD (also called geographic atrophy) and ``wet'', or neovascular, AMD. AMD = age-related macular degeneration; AREDS = Age-Related Eye Disease Study.}
    \label{fig:stage}
\end{figure}

We validate LTSA on the prognosis of two eye diseases, late AMD and POAG. AMD is the leading cause of legal blindness in developed countries,\cite{Kawasaki2010-gd, Eye_Disease_Study_Research_Group2000-ou} and the number of people with AMD worldwide is projected to reach 288 million by 2040.\cite{Wong2014-tn}
The disease is broadly classified into early, intermediate, and late stages. While early and even intermediate AMD are typically asymptomatic, late AMD is often associated with central vision loss, occurring in two forms: geographic atrophy and neovascular (``wet'') AMD (\textbf{Fig.~\ref{fig:stage}}).\cite{Ferris2013-ts}
To improve management plans for patients, it is important to understand the individualized risk of AMD progression. Patients with low risk may adopt a management plan that will minimize costs and burden of care on the healthcare system. In contrast, high-risk patients may receive a more aggressive management plan earlier in the disease progression in order to maintain vision as long as possible. POAG, too, is one of the leading causes of blindness in the United States and worldwide,\cite{Bourne2013-sm} as well as the leading cause among African Americans and Latinos.\cite{Sommer1991-ng, Jiang2018-ea}
The disease is projected to affect nearly 112 million people by 2040, over 5 million of whom may become bilaterally blind.\cite{Tham2014-ag}
Similar to AMD, POAG is asymptomatic until it reaches an advanced stage when visual field loss occurs. However, early detection and treatment can avoid most blindness caused by POAG.\cite{Tatham2015-ux} For both late AMD and POAG, accurately identifying high-risk patients as early as possible is critical to clinical decision-making, helping inform management, treatment planning, or patient monitoring.

To evaluate LTSA, we performed extensive experiments comparing our proposed method to a single-image baseline, which only uses the single last available image. This study leveraged two large, longitudinal imaging datasets: 49,452 images from 4,274 participants from the Age-Related Eye Disease Study (AREDS) for late AMD prognosis and 30,932 images from 1,597 participants from the Ocular Hypertension Treatment Study (OHTS) for POAG prognosis. As measured by the time-dependent concordance index, LTSA demonstrates consistently superior discrimination of disease risk on both late AMD and POAG prognosis. LTSA significantly outperforms the single-image baseline on 37 out of 40 head-to-head comparisons across a wide variety of prediction, or ``landmark'' times, and time horizons. Our results also strongly suggest the benefit of longitudinal image modeling for prognosis, where incorporating prior imaging enhances disease prognosis. Further, since LTSA leverages a temporal attention mechanism over the sequence of images, analysis of the learned attention weights uncovers which visits contribute most to prognosis. Indeed, the most recent visit is usually the single most important; however, we are able to characterize the relationship between time since the final examination and the relative influence of each exam on the predicted prognosis, which may have real-world, real-time implications for ophthalmologists making assessments of risk and prognosis.

Beyond the improved discriminative performance of our proposed method, this study offers a potential answer to the growing demand for \emph{dynamic} and \emph{explainable} prognoses for eye diseases. LTSA can enhance our understanding of temporal image patterns contributing to eye disease progression, serving to demystify ``black-box'' deep learning models. Such clarity could potentially promote greater utilization and trust of deep survival analysis models among ophthalmologists, bridging the gap between technical innovation and clinical practice.

\begin{figure}[htbp]
    \centering
    \includegraphics[width=1\linewidth]{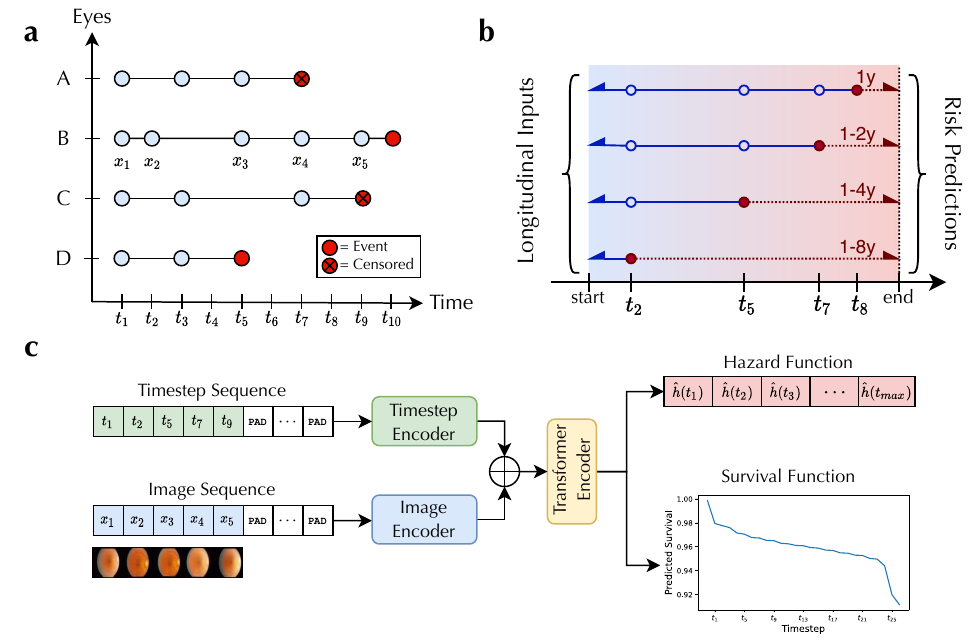}
    \caption{\textbf{Overview of proposed longitudinal survival analysis approach.} In longitudinal medical imaging, patients undergo repeated imaging over long periods of time at irregular intervals \textbf{(a)}. Rather than predict the presence of disease at the time of imaging, our method leverages a patient's longitudinal imaging history to forecast the \emph{future} risk of developing disease through a survival analysis framework \textbf{(b)}. Our approach represents the collection of fundus images for an eye over time as a \emph{sequence} fit for modeling with Transformers. To accommodate large, irregular intervals between consecutive visits, a temporal positional encoder fuses this information with the image embeddings from each visit. A Transformer encoder then employs causal temporal attention over the sequence, only attending to prior visits. The entire model is optimized end-to-end to predict the time-varying hazard function for each unique sequence of consecutive visits. From the hazard function, we compute eye-specific survival curves, allowing for dynamic eye disease risk prognosis evaluated through the framework of longitudinal survival analysis \textbf{(c)}.}
    \label{fig:overview}
\end{figure}

\section{Results}\label{results}

\subsection{Development and evaluation of LTSA}\label{development-and-evaluation-of-ltsa}

Our proposed LTSA is trained on sequences of consecutive fundus images to directly predict the time-varying hazard distribution, allowing for disease prognosis through a survival analysis framework (\textbf{Fig. \ref{fig:overview}}). Instead of standard positional encoding, a temporal positional encoder accounts for irregularly spaced imaging by embedding the time of each visit and fusing this information with the associated image embedding. Then, a Transformer encoder performs causal, temporal attention over the sequence of longitudinal images before a survival layer predicts the sequence-specific hazard function.

LTSA was trained and evaluated using longitudinal fundus imaging and time-to-event data from the Age-Related Eye Disease Study (AREDS) and Ocular Hypertension Treatment Study (OHTS). The AREDS data consisted of 49,592 images from 4,274 unique patients and 7,818 unique eyes, and the OHTS data consisted of 30,932 images from 1,597 patients and 3,188 eyes (see ``Methods'' for further details). In the AREDS dataset, eyes underwent an average of 6.34 visits over the course of 6.47 years, with a minimum of 6 months between visits. Approximately 12.2\% of eyes developed late AMD (87.8\% censoring rate) with a mean time to event of 5.27 years. For OHTS, eyes were examined an average of 9.70 times over 9.20 years with a minimum of 1 year between visits. Approximately 11.9\% of eyes developed glaucoma (88.1\% censored) in an average of 7.22 years. For model development and evaluation, each dataset was then randomly partitioned into training (70\%), validation (10\%), and test (20\%) sets at the patient level.

To account for censoring and time-varying inputs and outputs, we assess the prognostic ability of our models with a time-dependent concordance, denoted $C(t,\Delta t)$. This metric measures the ability to accurately rank pairs of eyes by risk (e.g., predicting higher risk for eyes that will develop disease sooner) for a given \emph{prediction time} $t$ (time of last visit) and \emph{evaluation time} $\Delta t$ (time horizon into the future over which we assess risk). We compare the performance of LTSA with a single-image baseline that only uses the most recent available image. Models are evaluated by $C(t,\Delta t)$ across 20 combinations of prediction times $t \in \{1,2,3,5,8\}$ and evaluation times ${\Delta}t \in \{1,2,5,8\}$, where time is measured in years. Finally, evaluation is performed at the eye, not patient, level since each eye can have its own unique disease status.

\subsection{Validation of LSTA on late AMD risk prognosis}\label{validation-of-lsta-on-late-amd-risk-prognosis}

\begin{figure}[!ht]
    \centering
    \includegraphics[width=\linewidth]{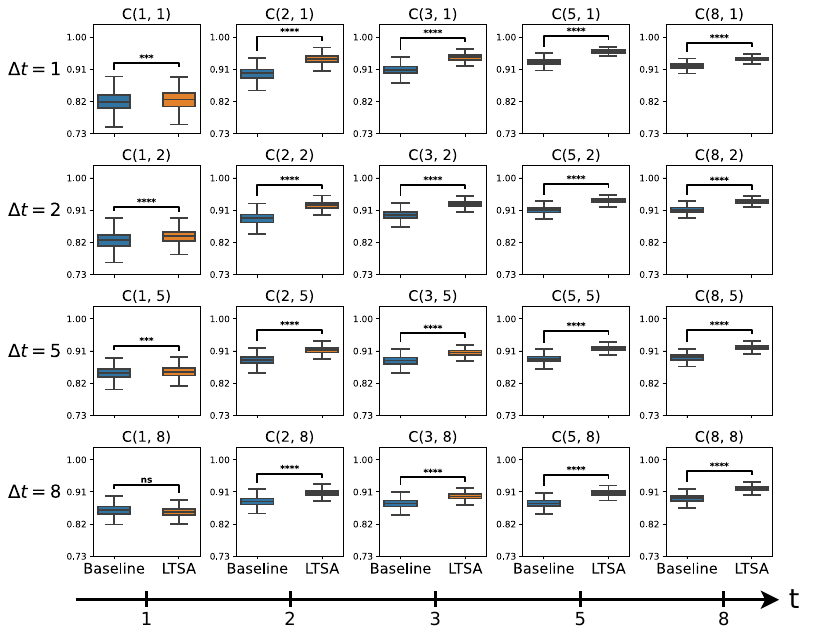}
    \caption{\textbf{Late AMD prognosis results.} Time-dependent concordance index $C(t,{\Delta}t)$ for various values of prediction time $t$ and evaluation time $\Delta t$ comparing the single-image baseline model (blue) to LTSA, which incorporates all prior visits (orange). Box plots depict the values computed from 1,000 bootstrap samples of the test set (center line = median, box = IQR, whiskers = 1.5x the IQR from the box). Significance levels are determined from Bonferroni-adjusted $P$-values as follows: **** = $P \leq 0.0001$, *** = $P \leq 0.001$, ** = $P \leq 0.01$, * = $P \leq 0.05$, ns = no significant difference. AMD = age-related macular degeneration; IQR = interquartile range.}
    \label{fig:late}
\end{figure}

\textbf{Fig. \ref{fig:late}} reveals that LTSA significantly outperformed the single-image baseline for late AMD prognosis in 19 out of 20 combinations of prediction time $t$ and evaluation time ${\Delta}t$. Across all prediction and evaluation times, the single-image baseline achieved a mean time-dependent concordance index of 0.884 (95\% CI: [0.861, 0.906]), while LTSA reached 0.907 (95\% CI: [0.890, 0.924]). Full numerical results for late AMD prognosis can be found in \textbf{Supplementary Table \ref{tab:full_amd}}. For $t = 1$, the baseline and LTSA were comparable, given that often only one image is available after a year of observation. For example, the baseline produced a $C(1,1)$ of 0.818 (95\% CI: [0.771, 0.862]) and $C(1,2)$ of 0.825 (95\% CI: [0.787, 0.861]), while LTSA produced a $C(1,1)$ of 0.823 (95\% CI: [0.775, 0.863]) and $C(1,2)$ of 0.836 (95\% CI: [0.802, 0.868]). However, for prediction times beyond one year -- when LTSA would have access to more prior imaging -- LTSA dramatically outperformed the single-image baseline in all 16 out of 16 comparisons ($P\leq 0.0001$ for each test). For example, considering 2-year late AMD risk prediction, $C(2,2)$ was 0.886 (95\% CI: [0.858, 0.911]) for the baseline vs. 0.923 (95\% CI: [0.903, 0.940]) for LTSA, $C(3,2)$ was 0.896 (95\% CI: [0.873, 0.917]) for the baseline vs. 0.927 (95\% CI: [0.911, 0.942]) for LTSA, and $C(5,2)$ was 0.910 (95\% CI: [0.894, 0.926]) for the baseline vs. 0.936 (95\% CI; [0.924, 0.947]) for LTSA. Similar trends were observed for 1-, 5-, and 8-year late AMD risk prognosis as well. Additional late AMD prognosis results by time-dependent Brier score can be seen in \textbf{Supplementary Figure \ref{fig:Auxiliary amd}}.

\subsection{Validation of LTSA on POAG risk prognosis}

\begin{figure}
    \centering
    \includegraphics[width=\linewidth]{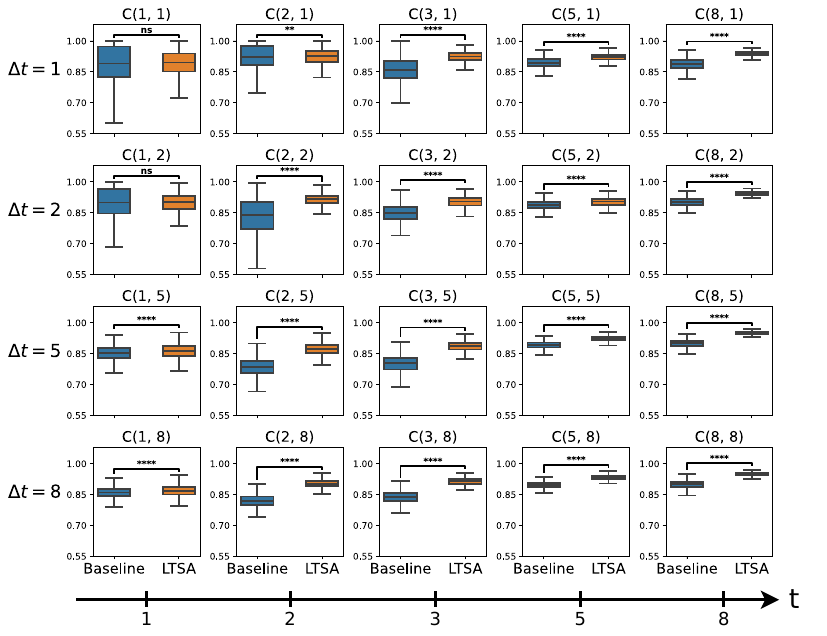}
    \caption{\textbf{POAG prognosis results.} Time-dependent concordance index $C(t,\Delta t)$ for various values of prediction time $t$ and evaluation time $\Delta t$ comparing the single-image baseline model (blue) to LTSA, which incorporates all prior visits (orange). Box plots depict the values computed from 1,000 bootstrapped samples of the test set (center line = median, box = IQR, whiskers = 1.5x the IQR from the box). Significance levels are determined from Bonferroni-adjusted $P$-values as follows: **** = $P \leq 0.0001$, *** = $P \leq 0.001$, ** = $P \leq 0.01$, * = $P \leq 0.05$, ns = no significant difference. IQR = interquartile range; POAG = primary open-angle glaucoma.}
    \label{fig:poag}
\end{figure}

\textbf{Fig. \ref{fig:poag}} shows that LTSA significantly outperformed the baseline on POAG prognosis for 18 out of 20 combinations of $t$ and ${\Delta}t$, as measured by the time-dependent concordance index. While the baseline reached an overall mean time-dependent concordance index of 0.866 (95\% CI: [0.795, 0.925]), LTSA reached 0.911 (95\% CI: [0.869, 0.948]). Full numerical results for POAG prognosis can be found in \textbf{Supplementary Table \ref{tab:full_poag}}. Once again, the performance gap between LTSA and the single-image baseline widened as more prior images became available; as prediction time $t$ increased, the number of significant improvements (and the magnitude of these improvements) of LTSA over the baseline was nondecreasing. LTSA also demonstrated an advantage in long-term POAG prognosis, even with early prediction times. For example, LTSA significantly outperformed the baseline for 5- and 8-year prognosis across \emph{all} prediction times ($P \leq 0.0001$ for all 10 comparisons). While LTSA provided a small but significant boost over the baseline for 5-year prognosis from year 1 -- $C(1,5)$ of 0.852 (95\% CI: [0.785, 0.910]) for the baseline vs. 0.861 (95\% CI: [0.800, 0.920], $P \leq 0.001$) for LTSA -- this gap only widened with increasing prediction time: $C(3,5)$ was 0.801 (95\% CI: [0.732, 0.865]) for the baseline vs. 0.885 (95\% CI: [0.846, 0.920]) for LTSA, and $C(8,5)$ was 0.899 (95\% CI: [0.863, 0.928]) for the baseline vs. 0.950 (95\% CI: [0.936, 0.962]) for LTSA. The same pattern could be observed for an 8-year POAG risk prognosis across the range of longitudinal prediction times. Auxiliary POAG prognosis results by time-dependent Brier score can be found in \textbf{Supplementary Figure \ref{fig:Auxiliary poag}}.

\subsection{Effect of longitudinal modeling on prognosis}\label{effect-of-longitudinal-modeling-on-prognosis}

\begin{figure}
    \centering
    \includegraphics[width=\linewidth]{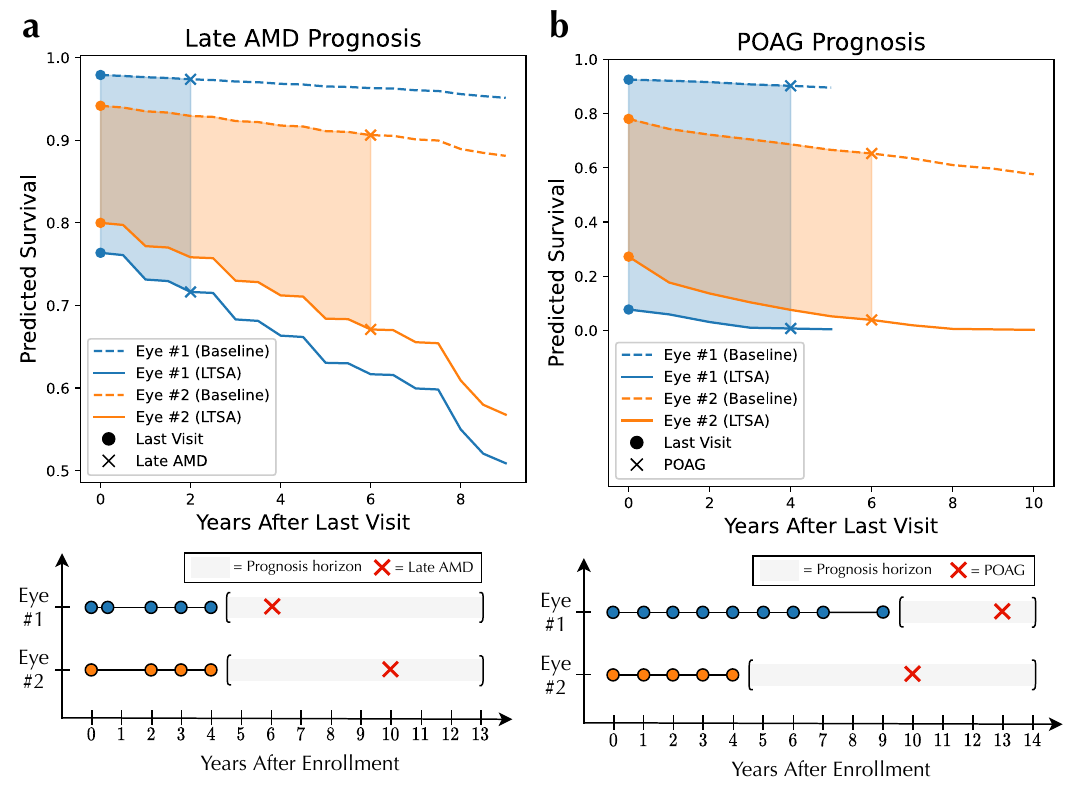}
    \caption{\textbf{Longitudinal modeling better captures eye disease risk.} Predicted survival curves comparing the baseline model (only using last available visit) and our longitudinal model (using all available visits) prognoses for two unique eyes in the AREDS test set \textbf{(a)} and two unique eyes in the OHTS test set \textbf{(b)}. Visualizations below each panel depict the longitudinal visit times, event times, and prognosis horizons for each eye in panels \textbf{a} and \textbf{b}, respectively. The longitudinal model not only correctly predicts higher risk (lower survival) than the baseline for each eye, but also correctly ranks the two eyes in terms of risk in accordance with how rapidly the eye will develop the disease. AMD = age-related macular degeneration. AREDS = Age-Related Eye Disease Study; OHTS = Ocular Hypertension Treatment Study.}
    \label{fig:Longitudinal}
\end{figure}

Based on the predicted time-varying hazard probabilities, LTSA can be used to dynamically construct eye-specific survival curves beginning from any time of interest. \textbf{Fig. \ref{fig:Longitudinal}a} depicts survival curves for two unique eyes in the AREDS test set, comparing the predicted survival trajectories of the single-image baseline (dashed line) to those of LTSA (solid line). Eye \#1 (blue) and eye \#2 (orange) both last underwent imaging 4 years from enrollment, though eye \#1 developed late AMD 2 years later and eye \#2 developed late AMD 6 years later. For both eyes, LTSA correctly predicts lower survival (higher risk) than the baseline, consistent with the fact that both eyes would go on to develop the disease. Additionally, LTSA correctly \emph{ranks} the eyes with respect to risk, while the single-image baseline does not -- that is, LTSA assigning lower survival probabilities to eye \#1 than eye \#2 is consistent with the fact that eye \#1 will go on to develop late AMD 4 years sooner than eye \#2. Similarly, \textbf{Fig. \ref{fig:Longitudinal}b} depicts an analogous pair of eyes from the OHTS test set, with predicted survival curves from LTSA and the single-image baseline. Eye \#1 (blue) was last observed during year 9, developing POAG 4 years later, while eye \#2 (orange) was last examined during year 4, developing POAG 6 years later. Here, the same pattern can be observed, where LTSA delivers a more accurate risk assessment than the baseline for both eyes. Notably, LTSA also properly ranks the two eyes according to glaucoma risk, a feat that the baseline does not. Conditional survival plots for these cases can be found in \textbf{Supplementary Figure \ref{fig:Auxiliary Longitudinal}}.

\subsection{Temporal attention analysis}\label{temporal-attention-analysis}

\begin{figure}
    \centering
    \includegraphics[width=\linewidth]{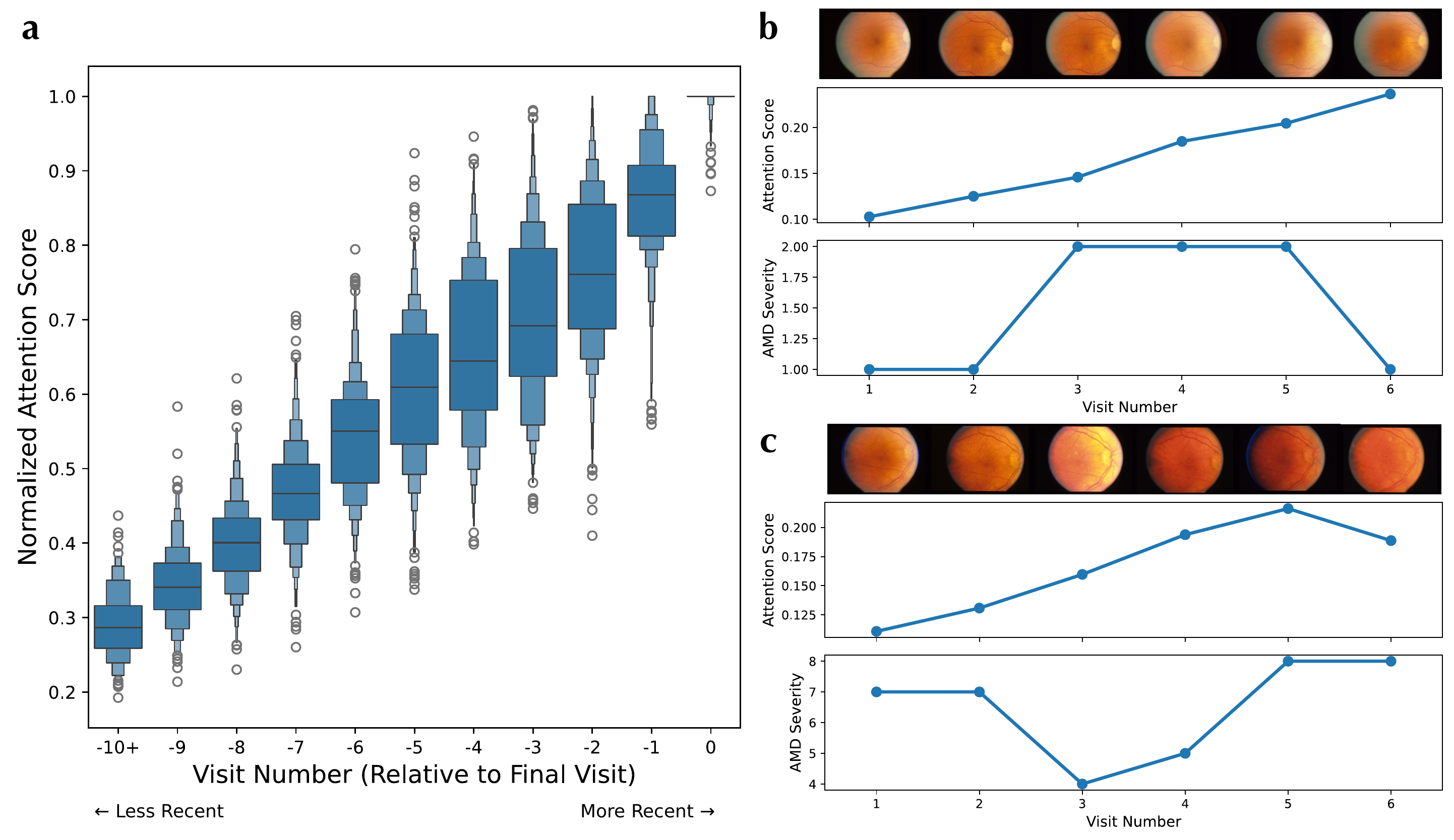}
    \caption{\textbf{Temporal attention analysis reveals which images contribute most to late AMD risk prognosis.} Enhanced box plot of normalized attention scores for each AREDS test set image grouped by visit number from least to most recent \textbf{(a)}. Attention scores and AMD severity scores for a sequence of visits from a healthy eye with typical attention patterns; the more recent visits are more influential than the earlier visits \textbf{(b)}. Attention scores and AMD severity scores for a sequence of visits from an eye that developed late AMD with atypical attention scores; here, the second-to-last image received the greatest attention weight, corresponding with a jump in AMD severity from 5 to 8 \textbf{(c)}. Attention scores are normalized such that the maximum score in each eye's sequence of images is 1 to account for variable sequence length. Images from 10 or more visits before the final visit are binned together to aid visualization. AMD = age-related macular degeneration.}
    \label{fig:Temporal}
\end{figure}

Since LTSA leverages a causal attention mechanism to process longitudinal imaging, the learned attention weights can reveal which visits are most influential to the model's disease prognosis. In support of common clinical practice and knowledge, temporal attention analysis suggests that, in the aggregate, more recent imaging is more important for late AMD risk prognosis (\textbf{Fig. \ref{fig:Temporal}a}). Across all unique eyes in the AREDS test set, the \emph{last} available image was given the highest attention score in nearly 96\% of cases. However, we find that LTSA still attends to prior imaging in a monotonically time-decaying fashion; the median normalized attention score -- relative to the maximum attention score in each sequence -- was 1 (most important) for the last visit, 0.864 (86.4\% as important) for the second-to-last visit, 0.812 (81.2\% as important) for the third-to-last visit, etc. with a strong negative linear association ($r = - 0.912$). Alongside the quantitative results demonstrating the benefit of longitudinal modeling, this attention analysis suggests that, while the most recent imaging is often the most important, prior imaging can still provide additional prognostic value.

While, on average, more recent imaging is more pertinent for risk prognosis, analysis of the learned attention weights for \emph{individual} sequences of eyes can illuminate abnormal cases worth further study. \textbf{Fig. \ref{fig:Temporal}b} shows the raw attention weights and ground-truth, ophthalmologist-determined AMD severity scores for a typical, healthy eye adhering to the overall pattern of attention scores -- the more recent the imaging, the higher the attention weight. However, \textbf{Fig. \ref{fig:Temporal}c} shows an atypical case, where the eye went on to develop late AMD and, more importantly, the \emph{second-to-last} visit received the greatest attention weight. For this eye, the second-to-last image was most influential to LTSA's prognosis, consistent with a jump in AMD severity from 5 to 8, potentially suggesting rapid progression of AMD.

\section{Discussion}\label{discussion}

In this work, we presented a novel method for survival analysis from longitudinal imaging, LTSA, and validated our approach on two eye disease prognosis tasks. Both quantitative and qualitative analysis demonstrated clear superiority of LTSA over a single-image baseline -- representing standard clinical practice -- for both late AMD and glaucoma prognosis. LTSA outperformed the baseline on 19 out of 20 head-to-head comparisons with the baseline for late AMD prognosis and 18 out of 20 comparisons for POAG prognosis when evaluated by the time-dependent concordance index. Given the longitudinal survival analysis setting with time-varying inputs and outputs, evaluation was performed across a wide range of prediction times $t$ and evaluation times ${\Delta}t$. LTSA particularly shined over the single-image baseline for $t \geq 2$, at which point multiple longitudinal images are often available for a given eye; in such cases, LTSA's causal attention mechanism allows for rich representation learning of changes apparent in the longitudinal image measurements of an eye over time. Qualitative analysis of predicted survival trajectories also showed that LTSA can produce more accurate risk assessments than the baseline for a given eye and accurately rank eyes with respect to risk when the single-image baseline cannot.

Our results suggest that longitudinal modeling can improve eye disease risk prognosis, providing evidence that prior imaging can provide added prognostic value. While longitudinal analysis is known to be clinically valuable, particularly for glaucoma prognosis, it can be time-consuming to perform such comparative image-based analysis in clinical practice. A temporal attention analysis of the learned attention weights by LTSA revealed that the last visit is almost always ($\sim$96\% of the time) the most influential for prognosis. However, in the aggregate, we find that LTSA consistently attends to prior imaging in order to make risk predictions, with more distant imaging becoming less important in a linearly decreasing manner. The unique sequence-based representation of longitudinal medical imaging and repeated temporal attention operations of LTSA enable us to study and uncover the importance of prior imaging in the context of eye disease prognosis. Meanwhile, existing medical image classification techniques are neither able to process sequences of images over time nor quantify their relative contribution to the predicted outcome.

Since methods like DeepSurv\cite{Katzman2018-pm} and Cox-nnet\cite{Ching2018-vs} popularized the use of deep neural networks for survival analysis, there have been many recent applications of deep learning to survival analysis from medical imaging data.\cite{Wiegrebe2023-lp, Wolf2022-vw, Zhang2020-km, Chen2021-ba, Agarwal2021-oa}
However, these methods all operate on single images, neglecting the common clinical scenario of longitudinal imaging, where patients undergo repeated imaging measurements over time in order to monitor changes in disease status. In recent years, several deep learning methods have also been proposed modeling time-to-event outcomes from longitudinal data. However, these studies have been limited to \emph{tabular} longitudinal or time series data.\cite{Wiegrebe2023-lp, Lee2020-fc, Lin2022-ls, Gupta2019-is}
Unlike the select few related methods capable of survival analysis from high-dimensional longitudinal medical imaging data,\cite{Shu2021-cg, Kang2023-ad}
LTSA critically (i) is an \emph{end-to-end} method (i.e., no multi-stage training), (ii) uses a Transformer encoder with causal attention to process sequences of medical images, and (iii) leverages an auxiliary ``step-ahead'' prediction task, whereby the model is tasked to predict the image-derived features from the \emph{next} visit (see ``Methods'' for full description).

This study has certain limitations. First, LTSA was developed for discrete-time survival modeling, when certain applications may call for more fine-grained continuous-time survival estimates. While discretizing time can often serve as a simplifying assumption, we adopted a discrete-time model because the AREDS and OHTS datasets followed up with patients at discrete 6-month (AREDS) or 1-year (OHTS) minimum intervals. Second, the validation of LTSA was limited to the prognosis of two eye diseases. However, the method can be readily applied to any disease prognosis task for which one has longitudinal imaging and time-to-event data. For example, LTSA could be used to predict Alzheimer's conversion from longitudinal neuroimaging\cite{Mirabnahrazam2023-kh, Nakagawa2020-pe} or survival of cancer patients based on various medical imaging modalities.\cite{Ul_Haq2022-aq, Liu2022-tb}
Third, while late AMD and POAG progression is unique to each eye, future work may incorporate binocular imaging to predict patient-level risk of disease progression. Also, incorporating tabular demographic and risk factor information may further improve prognostic performance. While this study used a fixed set of hyperparameters to enable fair comparison across methods, further hyperparameter optimization could be employed, particularly to analyze the impact of the regularization term $\beta$ on downstream prognosis performance. Finally, this study represents a retrospective analysis of clinical trial data that, even with broad eligibility criteria, may not generalize well to real-world populations. To bridge the gap toward clinical translation, real-world clinical assessment is needed to determine whether patients can practically benefit from these predictions, and if the approach is clinically safe and efficient. This will also critically allow us to refine benchtop-derived artificial intelligence according to bedside clinical demands.

\section{Methods}\label{methods}

\subsection{Study cohorts}\label{study-cohorts}

In this study, we included two independent datasets (\textbf{Table \ref{tab:Description}}). These datasets are derived from large, population-based studies, and the research adhered to the principles outlined in the Declaration of Helsinki. In addition, all participants provided informed consent upon entry into the original studies. The study protocol was approved by the Institutional Review Board (IRB) at Weill Cornell Medicine.

\begin{table}
\caption{\textbf{Description of longitudinal eye disease datasets.}}
\label{tab:Description}
\begin{center}
\vspace{-2em}
\begin{threeparttable}
\begin{tabular}{lllll}
\toprule
& Total & Training & Validation & Test\\
\midrule
\emph{AREDS (Late AMD)} & & & & \\
\hspace{1em}Images, n & 49,592 & 34,399 & 5,129 & 10,064 \\
\hspace{1em}Patients, n & 4,274 & 2,991 & 427 & 856 \\
\hspace{1em}Eyes, n & 7,818 & 5,461 & 782 & 1,575 \\
\hspace{1em}Visits, mean (sd) & 6.34 (3.23) & 6.30 (3.23) & 6.56 (3.16) & 6.39 (3.27) \\
\hspace{1em}Years observed, mean (sd) & 6.47 (3.43) & 6.44 (3.44) & 6.63 (3.33) & 6.50 (3.46) \\
\hspace{1em}Censored cases, n (\%) & 6,862 (87.8) & 4,777 (87.5) & 695 (88.9) & 1,390 (88.3) \\
\hspace{1em}Years to disease, mean (sd) & 5.27 (3.18) & 5.27 (3.18) & 5.84 (3.19) & 4.97 (3.12) \\
\midrule
\emph{OHTS (POAG)} & & & & \\
\hspace{1em}Images, n & 30,932 & 21,525 & 3,166 & 6,241 \\
\hspace{1em}Patients, n & 1,597 & 1,117 & 160 & 320 \\
\hspace{1em}Eyes, n & 3,188 & 2,231 & 320 & 637 \\
\hspace{1em}Visits, mean (sd) & 9.70 (3.68) & 9.65 (3.71) & 9.89 (3.60) & 9.80 (3.61) \\
\hspace{1em}Years observed, mean (sd) & 9.20 (3.70) & 9.14 (3.74) & 9.41 (3.56) & 9.29 (3.61) \\
\hspace{1em}Censored cases, n (\%) & 2,830 (88.8) & 1,966 (88.1) & 289 (90.3) & 575 (90.3) \\
\hspace{1em}Years to disease, mean (sd) & 7.38 (3.52) & 7.22 (3.61) & 7.71 (3.22) & 7.94 (3.18) \\
\bottomrule
\end{tabular}
\begin{tablenotes}
Characteristics of AREDS and OHTS longitudinal imaging datasets for late AMD and POAG prognosis, respectively. The number of visits and years observed are reported per eye since each eye can possess a distinct disease status. The censoring rate is reported as a percentage of the number of unique eyes in the dataset, and the number of years to develop the disease is computed only from uncensored eyes. AMD = Age-related Macular Degeneration; AREDS = Age-Related Eye Disease Study; OHTS = Ocular Hypertension Treatment Study; POAG = Primary Open-Angle Glaucoma; sd = standard deviation.
\end{tablenotes}
\end{threeparttable}
\end{center}
\end{table}

\paragraph*{The Age-Related Eye Disease Study (AREDS) for late AMD prognosis.} AREDS was a clinical trial conducted from 1992-2001 across 11 retinal specialty clinics throughout the U.S. to study the risk factors for AMD and cataracts and the effect of certain dietary supplements on AMD progression.\cite{Ferris2005-qi}
The study followed 4,757 participants aged 55-80 at the time of enrollment for a median of 6.5 years; the inclusion criteria were broad, ranging from no AMD in either eye to late AMD in one eye. Certified technicians captured color fundus photography images at baseline and in 6-month-to-1-year follow-up periods using a standardized imaging protocol; however, adherence to this protocol was imperfect, meaning visits could occur at any year or half-year mark after enrollment. AMD severity grades from each visit were then determined by human expert graders at the University of Wisconsin Fundus Photograph Reading Center. While AREDS involved the collection of many different types of patient information, the data used in the present study included 66,060 fundus images, the time (\emph{with 6-month temporal resolution}) that each image was acquired, and the ophthalmologist-determined AMD severity score using the 9-step severity scale.\cite{Davis2005-ul}
Late AMD was defined as the presence of one or more neovascular AMD abnormalities or atrophic AMD with geographic atrophy; otherwise, the eye was deemed to be censored, since the true late AMD status could not be known. All images acquired during the final visit for a given eye were removed, as this visit was solely used to determine the time-to-event outcome. Removing the final visit from the study cohort also ensured that no images were presented with late AMD at the time of acquisition, forcing the model to truly forecast the \emph{future} risk of developing disease. The remaining images from the remaining 4,274 patients were then randomly split into training (70\%), validation (10\%), and test (20\%) sets at the patient level to prevent any potential data leakage. All eligible images were included, regardless of image quality, to maximize the size of the training set and to ensure robustness to variations in image quality at test time.

\paragraph*{The Ocular Hypertension Treatment Study (OHTS) to predict the onset of POAG}. OHTS was one of the largest longitudinal clinical trials for POAG, spanning 22 centers across 16 U.S. states.\cite{Kass2002-nk}
The study followed 1,636 participants aged 40-80 with other inclusion criteria, such as intraocular pressure between 24-32 mmHg in one eye and 21-32 mmHg in the other eye. Color fundus photography images were captured annually, and POAG status was determined at the Optic Disc Reading Center. Much like the AREDS dataset, though visits were scheduled annually, adherence to this protocol was not exact, meaning visits could occur at any year mark after patient enrollment. In brief, two masked, certified readers were tasked to independently detect optic disc deterioration. If the two readers disagreed, a third senior reader reviewed it in a masked fashion. In a quality control sample of 86 eyes, POAG diagnosis showed a test-retest agreement of $\kappa = 0.70$ (95\% CI: [0.55, 0.85]); more details of the reading center workflow have been described in Gorden et al.\cite{Gordon1999-kn}
For the present study, 37,399 fundus images, their acquisition times (\emph{with 1-year temporal resolution}), and POAG diagnoses were used from OHTS. As outlined above for AREDS, we also removed all images from the final visit for each eye in the OHTS data. Additionally, in rare cases where there were multiple images acquired during a visit, we only kept the first listed image in our final OHTS cohort for simplicity. The 30,932 images from 1,597 patients were then randomly partitioned into training (70\%), validation (10\%), and test (20\%) sets at the patient level. Like the AREDS data, no images were removed for image quality reasons to encourage robustness to variations in quality.

\subsection{Longitudinal survival analysis}\label{longitudinal-survival-analysis}

We approach disease prognosis through the lens of survival analysis, which aims to model a ``time-to-event'' outcome from potentially time-varying input features. We adopted a discrete-time survival model, given that imaging measurements were either acquired at intervals as short as 6 months (AREDS) or 1 year (OHTS), and we assumed uninformative right-censoring. The collection of longitudinal images for eye $i$ can be written
\begin{equation}
    X_{i}(t) = \{x_{i}(t_{i,j}): 0<t_{i,j} \leq t,\ j = 1,\ldots,J_{i} \}
\end{equation}
where $J_{i}$ is the number of longitudinal images acquired for eye $i$, $t_{i,j}$ is the time (in years) of the $j^{th}$ image measurement for eye $i$, and $x_{i}\left( t_{i,j} \right) \in R^{H \times W}$is the fundus image (of height $H$ and width $W$) acquired at time $t_{i,j}$. Similar to the formulation of DynamicDeepHit,\cite{Lee2020-fc} we distinguish between discrete time steps $j$ and actual elapsed times $t$ since images are acquired at irregular intervals and the number of images per eye, $J_{i}$, is variable. In other words, $X_{i}(t)$ represents the collection of longitudinal images of eye $i$ acquired up until time $t$; for shorthand, we use $X_{i}$ to denote the \emph{full} available sequence of longitudinal images for eye $i$ (i.e., $X_{i} = X_{i}(t_{J_{i}})$). For each $X_{i}$, we also have a time to event $\tau_{i}$, which is either the time at which the event occurred (e.g., eye developed late AMD -- denoted $c_{i} = 0$) or the censoring time (e.g., the patient was lost to follow-up or the study ended -- denoted $c_{i} = 1$).

The goal of deep survival analysis in longitudinal imaging is to approximate a function that links the time to event to our time-varying image measurements. A typical way to reason about the time to event is through the \emph{hazard function}
\begin{equation}
h( j | X_{i} ) = P( T = j|T \geq j, X_{i})
\end{equation}
the conditional probability that eye $i$ develops the disease at a discrete time step $j$, based on longitudinal measurements $X_{i}$, given that the true event time step is greater than or equal to $j$. From the hazard function, we can readily compute the \emph{survival function}
\begin{equation}
S( j | X_{i} ) = P( T > j|X_{i} ) = \prod_{s = 1}^{j}{(1 - h( s|X_{i} ))}
\end{equation}
the probability that the eye $i$ does \emph{not} develop the disease (``survives'') past the time step $j$.

Specifically, in this study, we train a neural network $f( \cdot )$ to directly map from a longitudinal imaging sequence to the discrete hazard distribution
\begin{equation}
f( X_{i} ) = \hat{h}(X_{i}) = \{\hat{h}( {s|X}_{i} ),s = 1,\ldots,J_{\max}\}
\end{equation}
where $J_{\max}$ is the total number of discrete time steps, typically chosen based on properties of the dataset, time-to-event task, and computational constraints. While hazards are computed for all time steps, including those that have already occurred before the time step $t$, our primary interest lies in the \emph{future} hazards (i.e., $s = J_{i} + 1,\ldots,J_{\max}$). As explained below, we mask out prior time steps to properly optimize and evaluate models. For models trained on AREDS data -- captured in 6-month intervals with a maximum observed follow-up time of 13 years -- we have set $J_{\max} = 27$. For models trained on OHTS data -- acquired in 1-year intervals with a maximum follow-up of 14 years -- we have set $J_{\max} = 15$.

\subsection{LTSA model}\label{ltsa-model}

\paragraph*{Input representation.} The input to LTSA consists of a collection of longitudinal images $X_{i} = \{ x_{i}(t_{i,j}), j = 1,\ldots,J_{i}\}$ and their ``visit times'' $\{ v_{i,j}, j = 1,\ldots,J_{i}\}$, denoting the time (in months since study enrollment) that image $j$ of eye $i$ was acquired. To handle variable-length sequences, we right-pad the sequence with zeros to the maximum observed sequence length $l$ in the dataset ($l = 14$ for both AREDS and OHTS) when necessary to produce a padded sequence $X_{i}^{*} \in R^{l \times 3 \times H \times W}$. Critically, these padded inputs will be masked during modeling, optimization, and evaluation as described in the following sections. Much like how sentences are represented as sequences of words in deep natural language processing (NLP),\cite{Vaswani2017-hz} we represent the longitudinal imaging of an eye as a time-varying sequence fit for modeling with Transformers. However, unlike words in a sentence, the longitudinal images in each sequence are not ``equally spaced,'' potentially years having passed between consecutive visits.

\paragraph*{Temporal positional encoder.} Transformers typically use positional encoding (PE)\cite{Vaswani2017-hz} to inform the model as to the \emph{order} of elements in an input sequence. This can be accomplished using a fixed sinusoidal PE that maps the position of an element in a sequence to a higher-dimensional representation fit for deep neural network modeling:
\begin{align}
    {\textrm{PE}(k)}_{k,2i} & = \sin\left( \frac{k}{10000^{2i/d}} \right)\\
    {\textrm{PE}(k)}_{k,2i + 1} &= \cos\left( \frac{k}{10000^{2i/d}} \right)
\end{align}
for $i = 0,...,d/2$, where $k \in Z_{\geq 0}$ represents the position of a given element in the input sequence. To account for long, irregular time periods between consecutive longitudinal images, we adapt traditional PE to directly embed the visit time $v$ (measured in months) via
\begin{align}
{\textrm{TE}(v)}_{v,2i} = \sin\left( \frac{v}{10000^{2i/d}} \right)\\
{\textrm{TE}(v)}_{v,2i + 1} = \cos\left( \frac{v}{10000^{2i/d}} \right)
\end{align}
for $\ i = 0,...,d/2$. After computing the timestep encoding for the entire sequence, this produces a temporal positional embedding $e_{time} \in R^{l \times d}$. This approach is similar to continuous positional encoding in Sriram et al.\cite{Sriram2021-ni} except that we use the absolute visit time rather than visit time relative to the final visit.

\paragraph*{Image encoder.} While the timestep encoder produces our time step embeddings $e_{time}$, an image encoder is separately used to learn visit-level image embeddings. To do so, the padded sequence of images $X_{i}^{*}$ is flattened along the batch dimension and fed into a 2D image encoder $f_{img}( \cdot )$ to produce image embeddings

\begin{equation}
    e_{img} = \{f_{img}(x_{i}), x_{i} \in X_{i}^{*}\} \in R^{l \times d}
\end{equation}

where $d$ is the dimensionality of the image embedding. In LTSA, $f_{img}(\cdot)$ is parameterized by a ResNet18\cite{He2015-zm} convolutional neural network, which maps each image to a $d = 512$-dimensional feature vector. However, in principle, $f_{img}(\cdot)$ can be parameterized with any 2D image encoder.

\paragraph*{Transformer modeling.} Following the practice of many Transformer networks,\cite{Vaswani2017-hz, Devlin2018-nc} we inject knowledge of visit time via elementwise addition of the time step embeddings and image embeddings: $e = e_{time} + e_{img}$. Our time-infused embeddings $e \in R^{l \times d}$ are then fed into a Transformer encoder that employs repeated self-attention operations to learn temporal associations across the sequence of longitudinal images for each eye. Unlike a typical Transformer encoder for NLP, where the model may learn associations between all words in a passage of text, a clinician can only rely on current and \emph{prior} imaging to form a diagnostic decision. For this reason, we adopt ``decoder-style'' \emph{causal} attention masking, where a diagonal mask is applied to the attention weight matrix, enforcing that the model only attends to current and prior visits in adherence to clinical reality. Additionally, a padding mask is applied, where features resulting from the zero-padded inputs do not contribute to the attention computations. The output of this Transformer $T( \cdot )$ is then
\begin{equation}
\tilde{e} = T(e) \in R^{l \times d}
\end{equation}

\paragraph*{Survival prediction.} After Transformer modeling, our embedding features $\tilde{e}$ are then used to directly predict the discrete-time hazard function. To achieve this, we use a simple fully-connected layer with $J_{\max}$ output neurons, followed by a sigmoid activation:
\begin{equation}
\hat{h}(\tilde{e}) = \sigma(\textrm{FC}(\textrm{Dropout}(\tilde{e}))) \in R^{l \times J_{\max}}
\end{equation}
where $\sigma( \cdot )$ is the sigmoid function and $\textrm{Dropout}( \cdot )$ is the regularization technique that randomly zeroes out a specified fraction of weights.\cite{Srivastava2014-kt} 
Since the fully-connected layer is applied in parallel to all $l$ elements of the sequence, the final output $\hat{h}(\tilde{e})$ represents the discrete hazard distributions for all $l$ subsequences of consecutive visits in the original sequence. That is, we obtain a full survival prediction based on the longitudinal history of \emph{every} visit. However, we are often only interested in the prediction based on the full longitudinal history for eye $i$, ${\hat{h}(\tilde{e})}_{J_{i}} \in R^{J_{\max}}$.

\paragraph*{Step-ahead feature prediction.} In addition to the primary task of predicting the hazard function, we also leverage an auxiliary prediction task, whereby we use the features from each subsequence of consecutive visits to directly predict the learned image embedding from the \emph{next} visit. Alongside survival modeling, this encourages the model to learn features from longitudinal imaging measurements that are predictive of future imaging. This approach has been shown to improve discriminative performance in related methods such as DynamicDeepHit\cite{Lee2020-fc} and TransformerJM\cite{Lin2022-ls}.

Since a future visit can occur any time after the most recent visit, this becomes a time-varying prediction problem. To inform the model as to the time period over which to predict future imaging features, we adopt a version of the temporal positional encoding explained above. Rather than embedding the visit time $v$, we embed the \emph{relative} time elapsed between the current and subsequent visit $r_{k}: = v_{k + 1} - v_{k},\ k = 1,...,l - 1$. This enables the model to flexibly control feature prediction in a time-dependent manner. Since the final discrete difference $r_{l}$ does not exist, we set it to 0 and mask it out as explained below.

Formally, we compute the predicted features via
\begin{equation}
\hat{x}(\tilde{e}) = \text{FC}(\text{Dropout}(\tilde{e} + \textrm{TE}(r))) \in R^{l \times d}
\end{equation}

Similar to the survival prediction outlined above, these ``step-ahead'' predictions are computed for every subsequence of consecutive visits in the original sequence. However, since a future image only exists for the first $J_{i} - 1$ subsequences, we mask out all other step-ahead predictions.

\paragraph*{Loss functions.} Models were trained to predict the discrete-time hazard distribution by optimizing a cross-entropy survival loss from Chen et al.:\cite{Chen2021-ba}
\begin{equation}
L_{surv} = (1 - \beta)L_{ce} + \beta L_{uncensored}
\end{equation}
where
\begin{equation}
L_{ce} = - c_{i}\log(\hat{S}(\tau_{i}|X_{i}))
\end{equation}
is the main cross-entropy-based survival term and
\begin{equation}
L_{uncensored} = - {(1 - c}_{i})\log(\hat{S}(\tau_{i} - 1|X_{i})) - {(1 - c}_{i})\log(\hat{h}(\tau_{i}|X_{i}))
\end{equation}
is a regularization term to provide additional weight to uncensored cases. We use $\beta = 0.15$ following the default value in the implementation of Chen et al.\cite{Chen2021-Multimodal}

The model was additionally trained to predict the image features corresponding to the \emph{next} visit in a longitudinal sequence by minimizing the mean squared error $L_{pred}$ between predicted step-ahead features $\hat{x}(\tilde{e})$ and the corresponding image embeddings $e_{img}$ during the same forward pass. As explained above, this loss is only computed for the first $J_{i} - 1$ valid subsequences, for which there exists a subsequent longitudinal image of eye $i$.

Finally, LTSA was trained by optimizing the sum of these two loss terms
\begin{equation}
L = L_{surv} + L_{pred}
\end{equation}

\subsection{Single-image baseline}\label{single-image-baseline}

The problem formulation for our single-image baseline, which only uses the last available image for modeling, is obtained by simply modifying the input as follows:
\begin{equation}
X_{i}(t) = x_{i}\left( \max_{j = 1,...,J_{i}}\left\{ t_{i,j}:0 < t_{i,j} \leq t \right\}\right)
\end{equation}

Now, $X_{i}(t)$ is no longer a \emph{collection} of images, but rather the single most recent available image for eye $i$ up until time $t$. Here, the shorthand $X_{i}$ would simply refer to the last image of eye $i$.

The baseline model consisted of an image encoder $f_{img}(\cdot)$, also parameterized by a ResNet18, trained and evaluated on the last available image for each eye. The model utilized the same survival output layer and was trained with the survival loss $L_{surv}$ only. In other words, this baseline lacked all longitudinal modeling components: visit times, sequence representation, Transformer modeling, and step-ahead prediction.

\subsection{Model evaluation}\label{model-evaluation}

To evaluate the prognostic ability of our models, we use a time-dependent concordance index
\begin{equation}
C(t,{\Delta}t) = P(\hat{R}(t + {\Delta}t\ |\ X_{i}(t)) > \hat{R}(t + {\Delta}t\ |\ X_{i^{'}}(t))\ |\ \tau_{i} < \tau_{i^{'}},\tau_{i} < t + {\Delta}t,c_{i} = 0 \vee c_{i^{'}} = 0)
\end{equation}
for a given \emph{prediction time} $t$ (when the prediction is made) and \emph{evaluation time} $\Delta t$ (period into the future over which we are assessing risk). This measures the proportion of ``concordant pairs'' of eyes, where the model predicts higher risk -- over the time window $(t,\ t + \Delta t\rbrack$ -- for the eye that develops the disease earlier (or lower risk for the eye that develops the disease later). Here, $\hat{R}(t + {\Delta}t\ |\ X_{i}(t))$ is a risk score representing the predicted probability of experiencing the event within $(t,\ t + \Delta t\rbrack$ based on longitudinal measurements of eye $i$ up until time $t$. Specifically, this risk score is calculated from the predicted survival probabilities via
\begin{equation}
\hat{R}(t + {\Delta}t\ |\ X_{i}(t)) = P(t < T \leq t + {\Delta}t\ |\ T\  > \ t)\  = \ \frac{\hat{S}(t) - \hat{S}(t + {\Delta}t)}{\hat{S}(t)}
\end{equation}

Since we are only interested in risk assessment from the prediction time over the specified evaluation time, this is equivalent to ``masking'' out risk predictions from irrelevant time steps.

Compared to the original concordance index,\cite{Harrell1982-yx} a standard measure of discriminative ability in survival analysis, this metric allows for dynamic, time-varying risk predictions over arbitrary time horizons of interest. This metric is very similar to the time-dependent concordance index used in Lee et al.\cite{Lee2020-fc}, except that we assess our model based on the predicted hazards (rather than ``hitting times''). We use this metric for a single-risk outcome.

Additional evaluation was performed by time-dependent Brier score to assess model calibration:
\begin{equation}
B(t,{\Delta}t) = \sum_{i = 1}^{N}{(I\left( \tau_{i} < t + {\Delta}t \right)(1 - c_{i}) - \hat{R}(t + {\Delta}t|X_{i}(t)))^{2}}
\end{equation}
where $I( \cdot )$ is the indicator function. This definition follows that of Lee et al.\cite{Lee2020-fc}

Models were evaluated across a range of prediction times $t \in \{ 1,2,3,5,8\}$ and evaluation times ${\Delta}t \in \{1,2,5,8\}$. While it is common to consider 2- and 5-year risk for late AMD\cite{Peng2020-pp, Lee2022-sq, Vitale2020-jf} and POAG,\cite{Li2022-uk, Lin2022-Multi}, we also include an 8-year risk assessment to showcase the long-range prognostic capabilities of LTSA.

\subsection{Implementation details}\label{implementation-details}

All models were implemented and trained with PyTorch v2.0.1\cite{Paszke2019-xm}. Before training, all AREDS and OHTS images were downsampled to 224 x 224 resolution with bilinear interpolation to accelerate data loading. After loading each image, the following data augmentations were applied, each with probability 0.5: random rotation, color jitter, Gaussian blur, and a random resized crop back to 224 x 224. Each image was then standardized with the channel-wise mean and standard deviation across all training set images. The image encoder $f_{img}( \cdot$) was an ImageNet-pretrained ResNet18, with weights made available through \emph{torchvision} v0.15.2 (\url{https://pytorch.org/vision/stable/index.html}). This architecture was chosen because it is lightweight and demonstrated sufficient performance compared to more sophisticated and memory-intensive architectures in preliminary experiments. The Transformer encoder of LTSA contained four Transformer layers, each with eight attention heads, a feature dimensionality of $d = 512$, ReLU activation, and dropout 0.25. The Transformer was trained from scratch with a diagonal causal attention mask prohibiting the model from attending to future elements of a sequence.

All classification heads (survival layer and step-ahead prediction layer) used a dropout of 0.25 on the incoming feature vectors. Both the baseline and LTSA were trained for a maximum of 50 epochs using early stopping with a ``patience'' of 10 epochs using a validation metric of mean time-dependent concordance index across all 20 combinations of $t$ and $\Delta t$; specifically, if the validation metric did not improve for 10 consecutive epochs, training was terminated and weights from the best-performing epoch were used for evaluation to prevent overfitting. Both models were trained with the Adam optimizer\cite{Kingma2014-ik} and initial learning rate $1 \times 10^{- 4}$ with a ``reduce on plateau'' scheduler that halved the learning rate whenever the validation metric did not improve for 3 consecutive epochs. Since LTSA was trained with a batch size of 32 (sequences of length 14 each), the single-image baseline used a batch size of 448 (images) to match the number of examples seen per minibatch for fair comparison.

\subsection{Statistical analysis}\label{statistical-analysis}

All performance metrics in this study are represented by the mean and 95\% confidence interval obtained by bootstrapping the test set at eye level. Specifically, 1,000 samples with replacements of the same size as the original test set were drawn, and nonparametric confidence intervals were obtained through the percentile method. All $P$-values were obtained by a one-sided Welch's $t$-test with the null hypothesis that the mean of the bootstrapped time-dependent concordance indices for LTSA exceeded that of the baseline. To control the family-wise error rate and account for multiple comparisons, we apply the Bonferroni correction\cite{Bland1995-bx} to all $P$-values by multiplying each raw $P$-value by 40, the number of performance comparisons made in this study. Significance levels were determined by the adjusted $P$-values as follows: ****=$P \leq 0.0001$, *** = $P \leq 0.001$, ** = $P \leq 0.01$, * = $P \leq 0.05$, ns = no significant difference. The enhanced box plot, or ``letter-value plot,'' seen in \textbf{Fig. \ref{fig:Longitudinal}}, adapts the traditional box plot more appropriately for long-tailed distributions.\cite{Hofmann2017-wx}

\textbf{Data availability}

The AREDS and OHTS data used in this study are available through the National Center for Biotechnology Information (NCBI) database of Genotypes and Phenotypes (dbGAP) through controlled access. AREDS data can be accessed from \url{https://www.ncbi.nlm.nih.gov/projects/gap/cgi-bin/study.cgi?study_id=phs000001.v3.p1}, and OHTS data can be accessed from \url{https://www.ncbi.nlm.nih.gov/projects/gap/cgi-bin/study.cgi?study_id=phs000240.v1.p1}.

\section*{Code availability}

The code repository for this work is available at \url{https://github.com/bionlplab/longitudinal_transformer_for_survival_analysis}.

\section*{Acknowledgment}

This work is supported by the National Eye Institute under Award No. R21EY035296 and the National Science Foundation under Award Nos. 2145640 and 2306556. It is also supported by the NIH Intramural Research Program, the National Library of Medicine, and the National Eye Institute.

\section*{Competing interests}

The authors declare no competing interests.

\section*{Author contributions}

Study concepts/study design, G.H., Z.W., Y.P.; manuscript drafting or manuscript revision for important intellectual content, all authors; approval of the final version of the submitted manuscript, all authors; agrees to ensure any questions related to the work are appropriately resolved, all authors; literature research, G.H., M.L., Z.W., Y.P.; experimental studies, G.H., M.L.; human evaluation, S.H.V., K.K.; data interpretation and statistical analysis, G.H., M.L., R.Z., Q.Y; and manuscript editing, all authors.

\bibliographystyle{medline}
\bibliography{ref}

\newpage
\appendix
\setcounter{table}{0}
\setcounter{figure}{0}
\renewcommand\figurename{Supplementary Figure} 
\renewcommand\tablename{Supplementary Table}

\newpage


\begin{table}[!hbpt]
    \caption{Full late AMD prognosis results.}
    \label{tab:full_amd}
    \centering
\begin{threeparttable}
    \begin{tabular}{clcccc}
\toprule
$t$ & & $\Delta t=1$ & $\Delta t=2$ & $\Delta t=5$ & $\Delta t=8$\\
\midrule
1 & Baseline & 0.818 (0.771, 0.862) & 0.825 (0.787, 0.861) & 0.848 (0.818, 0.875) & 0.858 (0.834, 0.881) \\
& LTSA & \textbf{0.823 (0.775, 0.863)} & \textbf{0.836 (0.802, 0.868)} & \textbf{0.851 (0.826, 0.874)} & \textbf{0.854 (0.832, 0.876)} \\
2 & Baseline & 0.897 (0.862, 0.927) & 0.886 (0.858, 0.911) & 0.884 (0.860, 0.907) & 0.883 (0.860, 0.904) \\
& LTSA & \textbf{0.938 (0.918, 0.956)} & \textbf{0.923 (0.903, 0.940)} & \textbf{0.913 (0.895, 0.930)} & \textbf{0.907 (0.891, 0.921)} \\
3 & Baseline & 0.908 (0.885, 0.929) & 0.896 (0.873, 0.917) & 0.883 (0.862, 0.903) & 0.877 (0.858, 0.897) \\
& LTSA & \textbf{0.943 (0.928, 0.957)} & \textbf{0.927 (0.911, 0.942)} & \textbf{0.905 (0.890, 0.920)} & \textbf{0.897 (0.882, 0.911)} \\
5 & Baseline & 0.930 (0.914, 0.944) & 0.910 (0.894, 0.926) & 0.888 (0.870, 0.905) & 0.878 (0.858, 0.896) \\
& LTSA & \textbf{0.959 (0.951, 0.967)} & \textbf{0.936 (0.924, 0.947)} & \textbf{0.917 (0.905, 0.929)} & \textbf{0.906 (0.893, 0.919)} \\
8 & Baseline & 0.919 (0.905, 0.931) & 0.911 (0.897, 0.925) & 0.892 (0.876, 0.908) & 0.892 (0.875, 0.909) \\
& LTSA & \textbf{0.939 (0.929, 0.948)} & \textbf{0.934 (0.924, 0.943)} & \textbf{0.920 (0.908, 0.931)} & \textbf{0.919 (0.908, 0.931)} \\
\bottomrule
    \end{tabular}
\begin{tablenotes}
Detailed quantitative late AMD prognosis results by time-dependent concordance index \(C(t,{\Delta}t)\) for various values of prediction time t and evaluation time \({\Delta}t\). Presented is the mean \(C(t,{\Delta}t)\) from 1,000 bootstrap samples of the test set, with 95\% bootstrapped confidence intervals in parentheses. Results from our proposed LTSA appear in boldface with a grey background, while the baseline results appear in standard font with a white background. LTSA = Longitudinal Transformer for Survival Analysis. AMD -- Age-related Macular Degeneration.
\end{tablenotes}
\end{threeparttable}
\end{table}

\newpage

\begin{table}[!hbpt]
    \caption{Full POAG prognosis results.}
    \label{tab:full_poag}
    \centering
\begin{threeparttable}
    \begin{tabular}{clcccc}
\toprule
$t$ & & $\Delta t=1$ & $\Delta t=2$ & $\Delta t=5$ & $\Delta t=8$\\
\midrule
1 & Baseline & 0.881 (0.693, 0.991) & 0.892 (0.751, 0.981) & 0.852 (0.785, 0.910) & 0.860 (0.815, 0.904) \\
& LTSA & \textbf{0.891 (0.764, 0.996)} & \textbf{0.898 (0.817, 0.972)} & \textbf{0.861 (0.800, 0.920)} & \textbf{0.867 (0.817, 0.911)} \\
2 & Baseline & 0.917 (0.803, 0.988) & 0.830 (0.660, 0.972) & 0.783 (0.706, 0.854) & 0.819 (0.765, 0.868) \\
& LTSA & \textbf{0.925 (0.859, 0.990)} & \textbf{0.913 (0.863, 0.957)} & \textbf{0.870 (0.821, 0.915)} & \textbf{0.902 (0.872, 0.931)} \\
3 & Baseline & 0.858 (0.746, 0.970) & 0.848 (0.777, 0.913) & 0.801 (0.732, 0.865) & 0.837 (0.789, 0.884) \\
& LTSA & \textbf{0.923 (0.879, 0.962)} & \textbf{0.902 (0.856, 0.945)} & \textbf{0.885 (0.846, 0.920)} & \textbf{0.914 (0.886, 0.939)} \\
5 & Baseline & 0.894 (0.852, 0.933) & 0.886 (0.848, 0.921) & 0.892 (0.863, 0.919) & 0.896 (0.871, 0.919) \\
& LTSA & \textbf{0.920 (0.893, 0.946)} & \textbf{0.903 (0.868, 0.934)} & \textbf{0.924 (0.901, 0.944)} & \textbf{0.933 (0.914, 0.952)} \\
8 & Baseline & 0.887 (0.842, 0.927) & 0.901 (0.865, 0.931) & 0.899 (0.863, 0.928) & 0.898 (0.863, 0.928) \\
& LTSA & \textbf{0.937 (0.918, 0.954)} & \textbf{0.944 (0.928, 0.958)} & \textbf{0.950 (0.936, 0.962)} & \textbf{0.949 (0.935, 0.962)} \\
\bottomrule
    \end{tabular}
\begin{tablenotes}
Detailed quantitative POAG prognosis results by time-dependent concordance index \(C(t,{\Delta}t)\) for various values of prediction time t and evaluation time \({\Delta}t\). Presented is the mean \(C(t,{\Delta}t)\) from 1,000 bootstrap samples of the test set, with 95\% bootstrapped confidence intervals in parentheses. Results from our proposed LTSA appear in boldface with a grey background, while the baseline results appear in standard font with a white background. LTSA = Longitudinal Transformer for Survival Analysis. AMD -- Age-related Macular Degeneration.
\end{tablenotes}
\end{threeparttable}
\end{table}

\newpage

\begin{figure}[!hbpt]
    \centering
    \includegraphics[width=\linewidth]{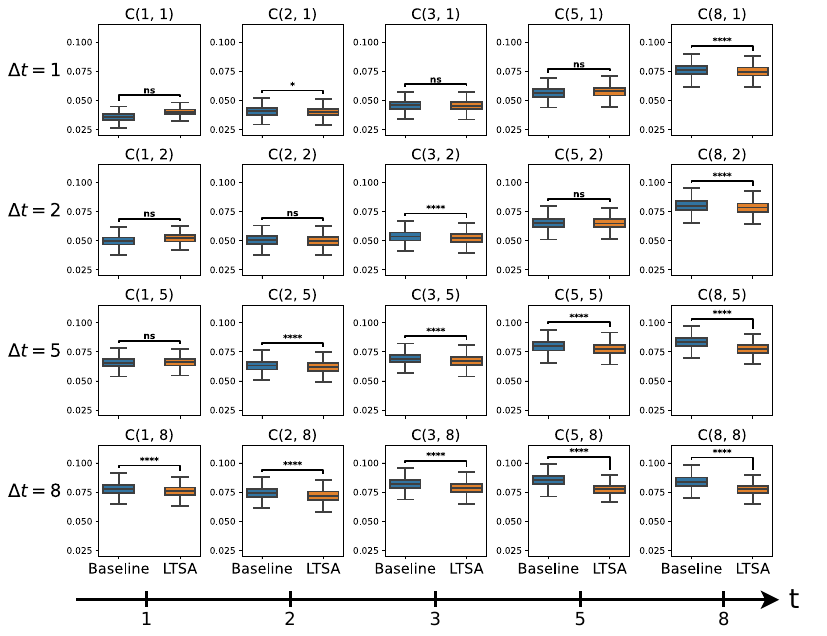}
    \caption{\textbf{Auxiliary late AMD prognosis results.} Time-dependent Brier score \(B(t,\ {\Delta}t)\) for various values of prediction time \emph{t} and evaluation time $\Delta t$ comparing the single-image baseline model (blue) to LTSA, which incorporates all prior visits (orange). Box plots depict the values computed from 1,000 bootstrap samples of the test set (center line = median, box = IQR, whiskers = 1.5x the IQR from the box). Significance levels are determined from Bonferroni-adjusted $P$-values as follows: **** = $P \leq 0.0001$, *** = $P \leq 0.001$, ** = $P \leq 0.01$, * = $P \leq 0.05$, ns = no significant difference. AMD = age-related macular degeneration; IQR = interquartile range.}
    \label{fig:Auxiliary amd}
\end{figure}

\newpage

\begin{figure}[!hbpt]
    \centering
    \includegraphics[width=\linewidth]{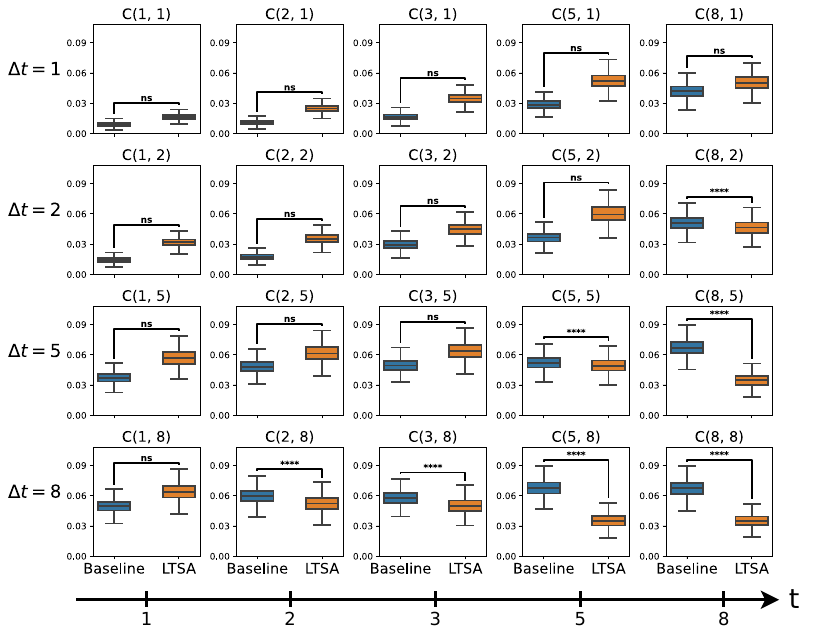}
    \caption{\textbf{Auxiliary POAG prognosis results.} Time-dependent Brier score \(B(t,\ \mathrm{\Delta}t)\) for various values of prediction time \emph{t} and evaluation time $\Delta t$ comparing the single-image baseline model (blue) to LTSA, which incorporates all prior visits (orange). Box plots depict the values computed from 1,000 bootstrap samples of the test set (center line = median, box = IQR, whiskers = 1.5x the IQR from the box). Significance levels are determined from Bonferroni-adjusted $P$-values as follows: **** = $P \leq 0.0001$, *** = $P \leq 0.001$, ** = $P \leq 0.01$, * = $P \leq 0.05$, ns = no significant difference. IQR = interquartile range; POAG = primary open-angle glaucoma.}
    \label{fig:Auxiliary poag}
\end{figure}

\newpage

\begin{figure}[!hbpt]
    \centering
    \includegraphics[width=\linewidth]{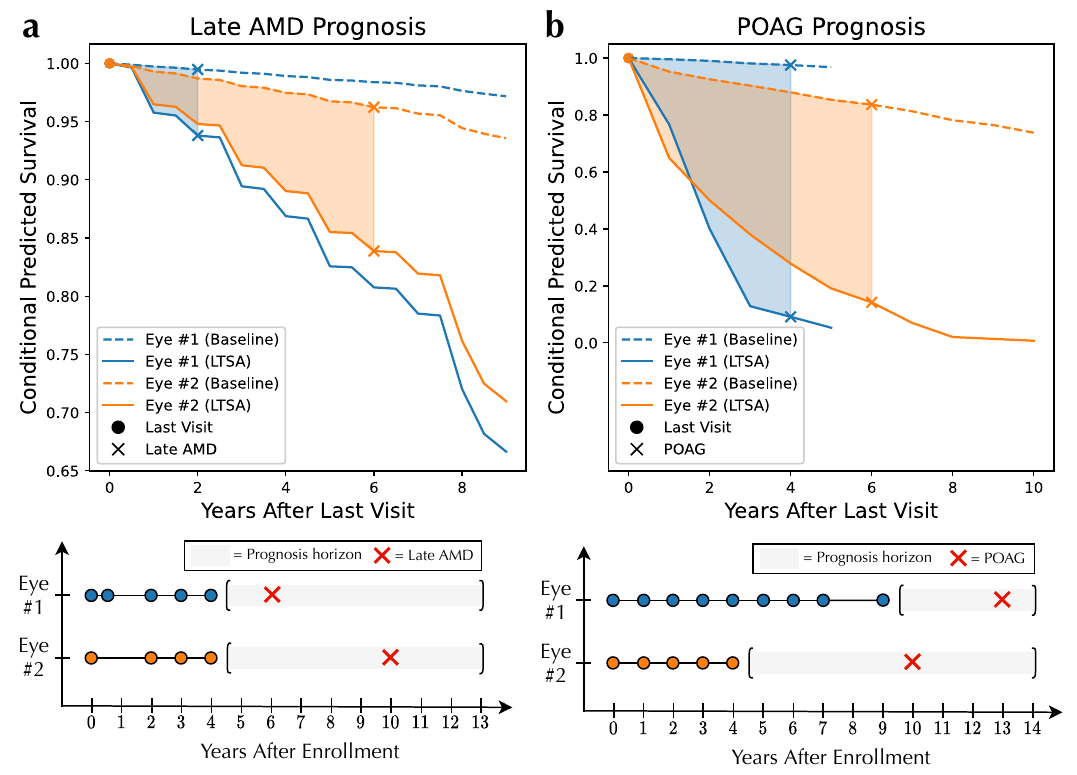}
    \caption{\textbf{Longitudinal modeling better captures eye disease risk.} Predicted conditional survival curves comparing the baseline model (only using last available visit) and our longitudinal model (using all available visits) prognoses for two unique eyes in the AREDS test set \textbf{(a)} and two unique eyes in the OHTS test set \textbf{(b)}. Visualizations below each panel depict the longitudinal visit times, event times, and prognosis horizons for each eye in panels \textbf{a} and \textbf{b}, respectively. AMD = age-related macular degeneration. AREDS = Age-Related Eye Disease Study; OHTS = Ocular Hypertension Treatment Study; POAG = primary open-angle glaucoma.}
    \label{fig:Auxiliary Longitudinal}
\end{figure}


\end{document}